# Time Series Comparisons in Deep Space Network

Kyongsik Yun[1], Rishi Verma[2], Umaa Rebbapragada[3]

*NASA Jet Propulsion Laboratory, California Institute of Technology, Pasadena, CA, 91109, USA*

The Deep Space Network (DSN) is NASA's international array of antennas that support interplanetary spacecraft missions. DSN provides radar and radio astronomy observations that enhance our understanding of the solar system and the larger universe. A track is a block of multi-dimensional time series from the beginning to end of DSN communication with the target spacecraft, containing thousands of monitor data items lasting several hours at a frequency of 0.2-1Hz. Monitor data on each track reports on the performance of specific spacecraft operations and the DSN itself. DSN is receiving signals from 32 spacecraft across the solar system. DSN has pressure to reduce costs while maintaining the quality of support for DSN mission users. DSN Link Control Operators (LCOs) need to simultaneously monitor multiple tracks and identify anomalies in real time. DSN has seen that as the number of missions increases, the data that needs to be processed increases over time. In this project, we look at the last 8 years of data for analysis. Any anomaly in the track indicates a problem with either the spacecraft, DSN equipment, or weather conditions. DSN operators typically write Discrepancy Reports (DR) for further analysis. It is recognized that it would be quite helpful to identify 10 similar historical tracks out of the huge database to quickly find/match anomalies. This tool has three functions: (1) identification of the top 10 similar historical tracks, (2) detection of anomalies compared to the reference normal track, and (3) comparison of statistical differences between two given tracks. The requirements for these features were confirmed by survey responses from 21 DSN operators and engineers. The preliminary machine learning model has shown promising performance (AUC=0.92). We plan to increase the number of data sets and perform additional testing to improve performance further before its planned integration into the track visualizer interface to assist DSN field operators and engineers.

## I. Nomenclature

*SS*    =  Similarity Score
*PC*    =  Pearson's Correlation
*ED*    =  Euclidean Distance
*DTW*  =  Dynamic Time Warping
*k*     =  distance calibration factor (i,e., 2~10)

## II. Introduction

Deep Space Network (DSN) was established in 1963 to provide the communications infrastructure for deep space missions [1]. DSN facilities span three locations on Earth to cover the entire universe as much as possible: Goldstone, California, USA; Madrid, Spain; and Canberra, Australia. Antenna arrays at these three locations support interplanetary spacecraft missions. The role of the DSN is to make radar and radio astronomy observations that advance our understanding of the solar system and the larger universe.

---

[1] Technologist, Autonomous Systems Division
[2] Software Systems Engineer, Mission Systems and Operations Division
[3] Data Scientist, Mission Systems and Operations Division

An example of streaming data from the DSN is shown in Fig 1. A "track" is a multidimensional time series block from the beginning to the end of a DSN communication with a target spacecraft, containing thousands of monitor data items. Monitoring data from each track provides information about specific spacecraft operations and the performance of the DSN itself. DSN is receiving signals from 32 spacecraft across the solar system.

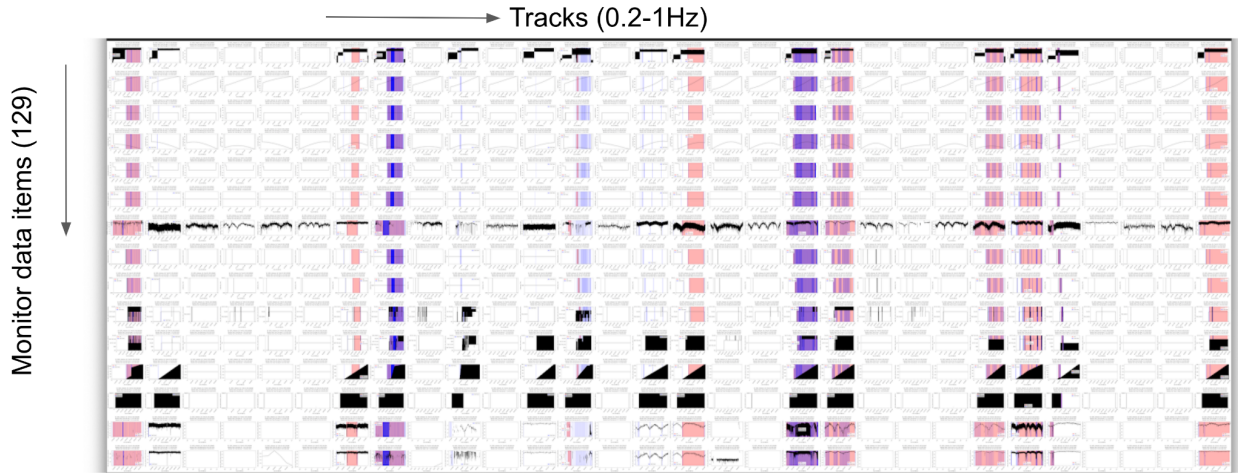

**Fig 1. Examples of multi-track and multi-monitor data items. Horizontal panels represent different tracks over time. Vertical panels represent different monitor data items. The first row is the monitor data item AGC_VOLTAGE (automatic gain control). Over time, AGC_VOLAGE sometimes exhibited unusual behavior (highlighted in red). This anomaly is also seen in other monitor data items.**

Historically, DSN operators worked 24 hours a day, 7 days a week with three shifts per day, and one operator per antenna. A new concept, Follow the Sun, launched on November 6, 2017, allowed each of the three DSN sites to operate the entire network during their day shift [2]. DSN is under pressure to reduce costs while maintaining the quality of support for DSN mission users. DSN operators must simultaneously monitor multiple tracks (up to 4 tracks) and identify anomalies in real time. DSN has found that as the number of missions increases, the amount of data it has to process increases over time. As of 2015, the maximum aggregate data capture rate was 150 Mbps [3], and the speed is expected to increase to up to 8000 Mbps by 2030 [4] as the number of missions is expected to increase from 32 to 80 [4].

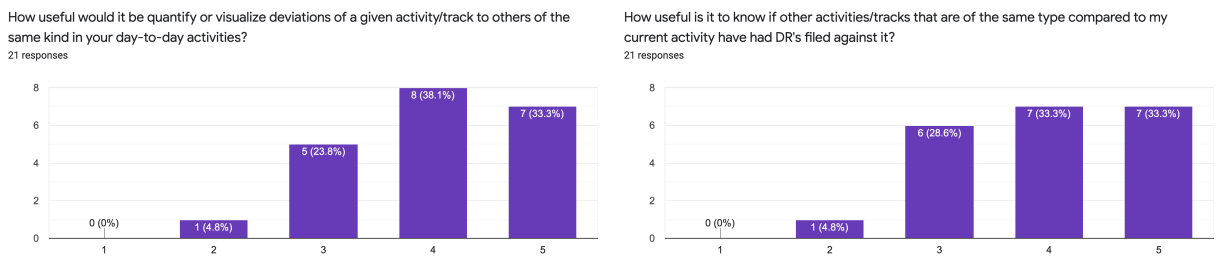

**Fig 2. DSN Link Control Operator (LCO) survey of useful analytical tools. A score of one is not at all useful, and a score of five is extremely useful. A Discrepancy Report (DR) will be submitted if there is any anomaly with the track. Finding similar historical tracks with DR has been found to be useful.**

We asked 21 DSN operators and engineers about useful analytical tools and found the track comparison tool to be very useful for operators' daily activities (Fig 2). 71.4% (15/21) of DSN operators said the track comparison tool would be useful, and 66.7% (14/21) of DSN operators said the track comparison tool would be particularly useful for abnormal tracks. DSN operators can find potential anomalies faster and more efficiently based on discrepancy reports from similar historical tracks.

## III. Track Comparisons Tool

The track comparison tool has three functions: (1) identifying the top 10 similar past tracks, (2) detecting anomalies compared to a reference normal track, and (3) comparing statistical differences between two given tracks. We can input one target track and compare similarity with past tracks of the same type (e.g., spacecraft, antenna) to produce an output of the top 10 similar tracks. We can also input two target tracks and compare their similarity scores to generate the output of similar or dissimilar classifications of the two tracks (Fig 3).

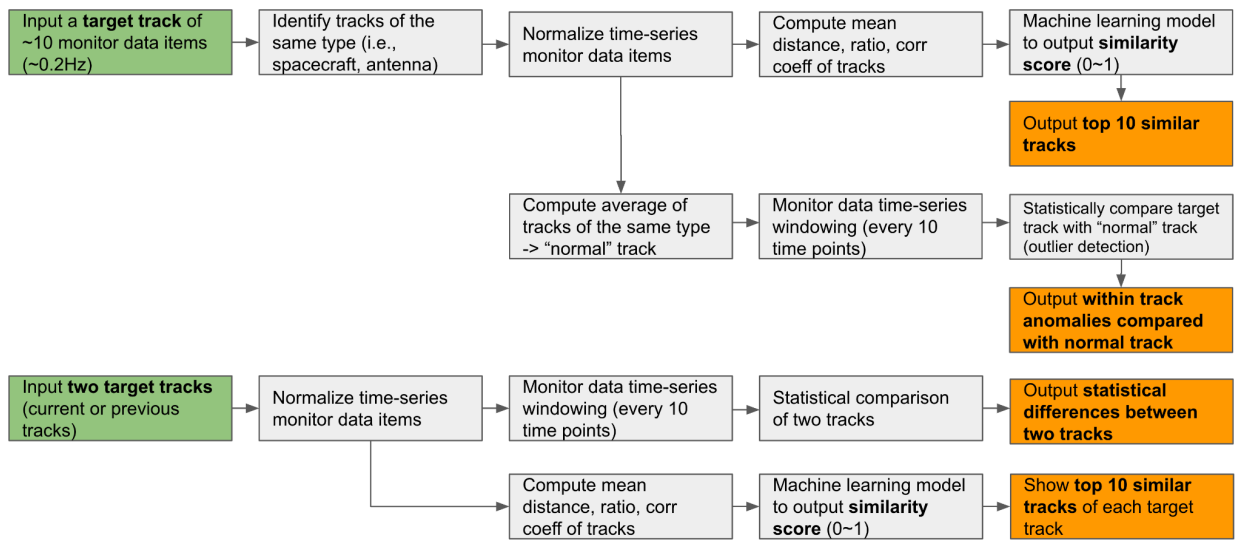

**Fig 3. Track comparison tool architecture diagram. Input data is green, output result is orange.**

We used three different time series comparison methods. First, the Euclidean (L2) distance was used. This is the simplest and most widely used method. But what if the time series are of different lengths? Dynamic Time Warping (DTW) can be used as a measure of elastic dissimilarity [5]. Euclidean distance does not accurately represent the distance between two temporally unsynchronized time series. DTW can solve the problem. Finally, the correlation coefficient was used. The ensemble method was then devised by combining the three methods.

$$SS = PC - (ED + DTW)/k \tag{1}$$

Correlation coefficients are best for DSN operators because they are intuitive and easy to understand. We aim to generate a similarity score that provides a holistic picture of the similarity of the two tracks. So what are the benefits of similarity scores as compared to Pearson's correlation? We found that similarity scores indicated greater contrast between high and low correlation tracks (T(18)=14.63, P<0.0001) than Pearson's correlations (T(18)=9.71, P<0.0001). This is due to the small correction for the similarity score in highly correlated tracks and the large correction to the similarity score in the low correlation tracks (Fig 4).

We also examined the relationship between Pearson's correlations and similarity scores across various antennas (DSS = 55, 56, 65) and monitoring data items. The same trend still holds in that small corrections occur on highly correlated tracks and large corrections occur on low correlation tracks.

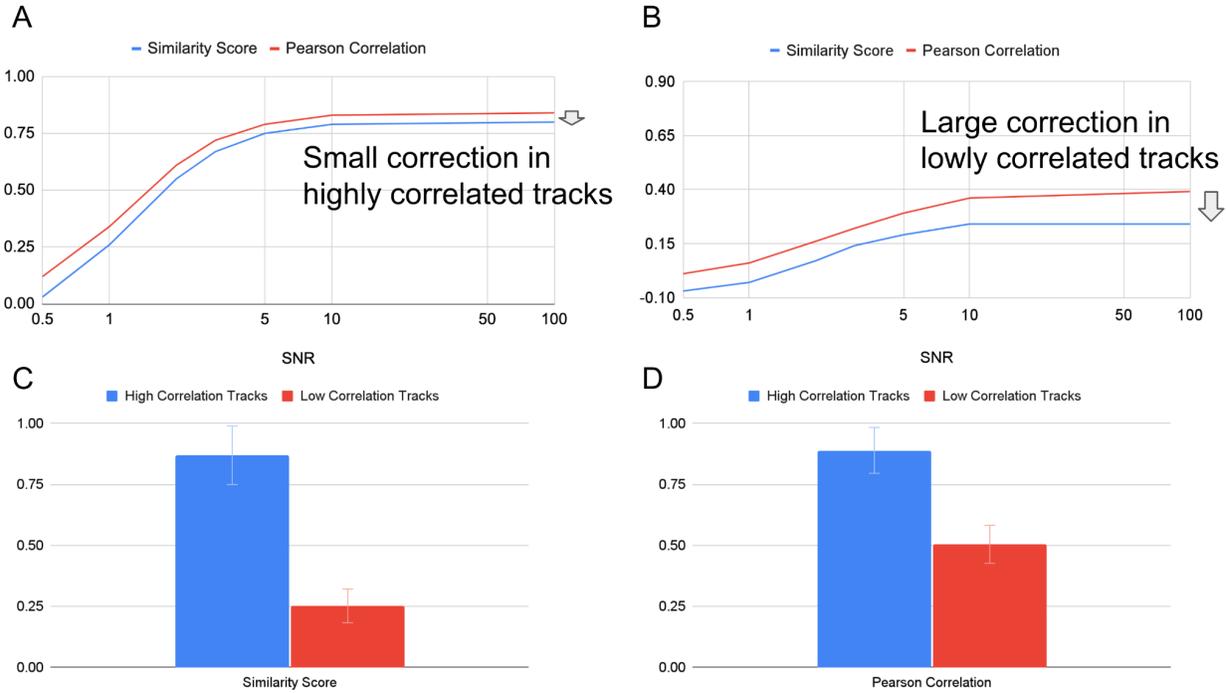

Fig 4. Similarity score validation for various signal-to-noise ratios (SNRs). (A) Pearson's correlation with similarity scores for different SNRs of highly correlated tracks. (B) Similarity scores and Pearson's correlations for different SNRs of low-correlated tracks. (C) Similarity scores of high and low correlation tracks. (D) Pearson correlation of high and low correlation tracks. Similarity scores showed greater contrast between high-correlation and low-correlation tracks than Pearson's correlations.

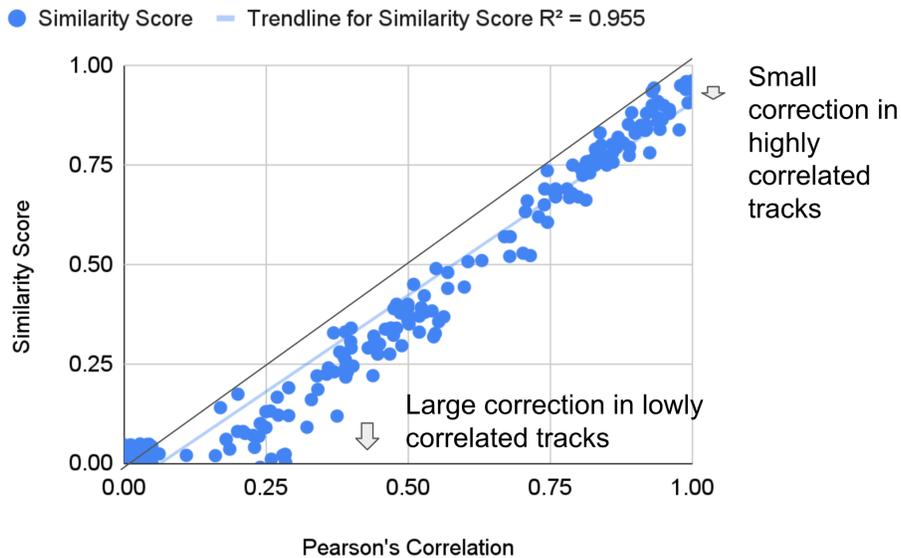

Fig 5. The relationship between Pearson's correlation and similarity scores. Similarity scores were modified from Pearson's correlations. The highly correlated tracks had small corrections and the low correlated tracks had large corrections, resulting in higher contrast in similarity scores (better classification) than Pearson's correlations.

We also found that the preliminary machine learning model showed promising performance (neural network AUC=0.92). We input multiple tracks (0.2 to 1 Hz with 10k to 20k time series data points per track) and monitor data items (7 items selected by DSN operator: carrier power, carrier system noise temperature, carrier track loop lock status, subcarrier track loop lock status, symbol rate, symbol track loop state, telemetry frame sync lock state). First, we identified the same type of track (spacecraft, antenna, communication type). Among the tracks of the same type, DSN operators manually identified 10 similar tracks and 10 dissimilar tracks (ground truths). The track comparison tool then normalizes the monitor data items for proper comparison and then calculates the Euclidean distance, DTW, and the correlation coefficient of the tracks. A classification model was then constructed using the three parameters (Euclidean distance, DTW, Pearson's correlation) with the seven monitor data items. The output target was the similar/dissimilar track classifications. We applied various machine learning models, such as Decision Tree (AUC=0.60), Logistic Regression (AUC=0.73), Naive Bayes (AUC=0.67), Linear Support Vector Machine (AUC=0.73), K-Nearest Neighbors (AUC=0.73), Random Forest (AUC=0.80) and feed-forward neural networks (AUC=0.92).

## IV. Time Series Data Compression

DSN time series data is stored in real time in Kafka and then in ElasticSearch [6]. We access ElasticSearch to extract historical tracks and Kafka for ongoing tracks. Extracting historical tracks in ElasticSearch results in significant data retrieval time delays (10-100 seconds), mainly due to complex queries involving nested queries to extract target tracks under certain conditions and large databases to search. ElasticSearch allows you to extract 10,000 time series data points per query, typically requiring 20-100 iterations of ElasticSearch to extract relevant tracks with the target conditions. The amount of data to extract per query is a bottleneck preventing real-time processing of track comparisons. Meanwhile, an ongoing track can be extracted from the Kafka streaming system in real time within 1ms (receiving streaming data) (Fig 6).

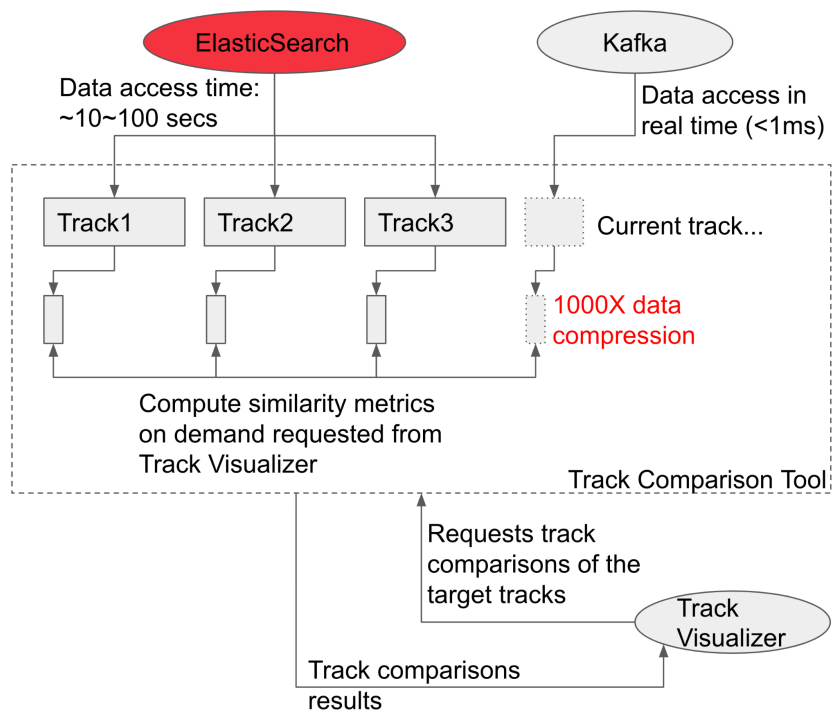

**Fig 6. Data flow in the DSN track comparison tool. ElasticSearch's data retrieval latency is more than 10 seconds, which requires retrieving data offline and compressing it by 1000X using piecewise polynomial approximation. Track Visualizer requests a track comparison to the Track Comparison Tool, and the system calculates a similarity metric on demand.**

In the real-time DSN track comparison use case, the time lag of tens to hundreds of seconds due to Elasticsearch data retrieval is not feasible. We need to access Elasticsearch track data offline, compress the time series data to at least 1000X to access the compressed database in real time, and compute similarity metrics on demand. We used a piecewise polynomial approximation to compress time series data [7–9]. Piecewise polynomial fitting is a popular data compression method that approximates a raw data stream with multiple polynomials. The polynomial coefficients corresponding to the best-fit curve can be calculated by the least squares method that minimizes the sum of the squared residuals between the observed and fitted values. We tested a variable number of hinge points from 5 to 30 and found that the approximation achieved a storage savings of 1000x while accurately modeling the original time series at more than 20 hinge points (Fig 7).

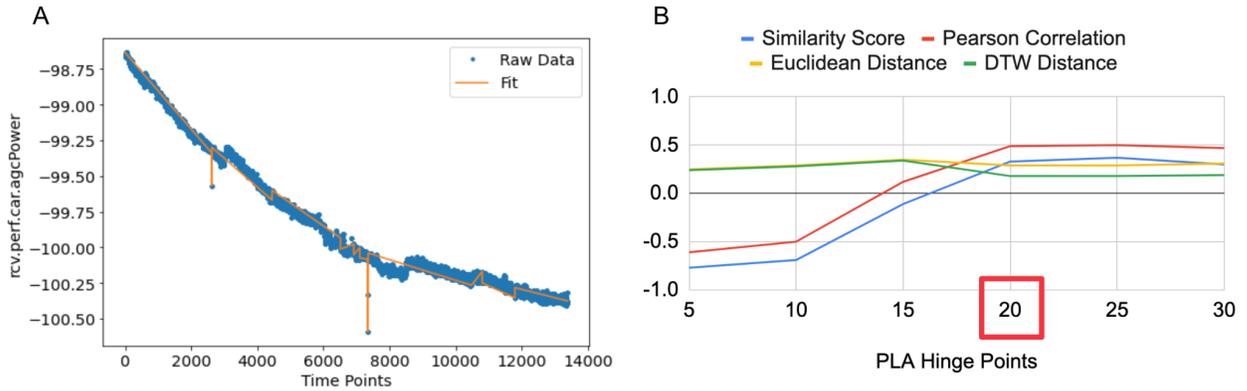

**Fig 7. Piecewise polynomial (linear) approximation for time series data compression (1000X). (A) Example monitor data item (AGC Power) raw data points (blue) and approximate results (orange). This approximation accurately captured both typical time series trends and anomalies (two spikes). (B) Similarity score, Pearson correlation, Euclidean distance, and DTW distance values at different numbers of hinge points. At more than 20 hinge points, all similarity metrics are saturated with the same value. This means that 20 hinge points are sufficient to accurately model the track time series.**

The next question will be the degree of accuracy degradation after data compression. We compared Pearson's Correlation and similarity score between raw and approximate (compressed) data (Fig 8). We found that data compression approximates the original correlation and similarity scores well ($R^2$=0.998) in similar (R>0.5) and dissimilar (R<-0.5) tracks (red boxes in Fig 8(A)). Compression error increases in low-correlated tracks (R>-0.3 or R<0.3) (orange box in Fig 8(A)). Compared with Pearson's correlation, the similarity score increased the contrast of the low correlation track (green box in Fig 8(B)), which is consistent with the results of Fig 5. that the contrast is greater in low correlation tracks.

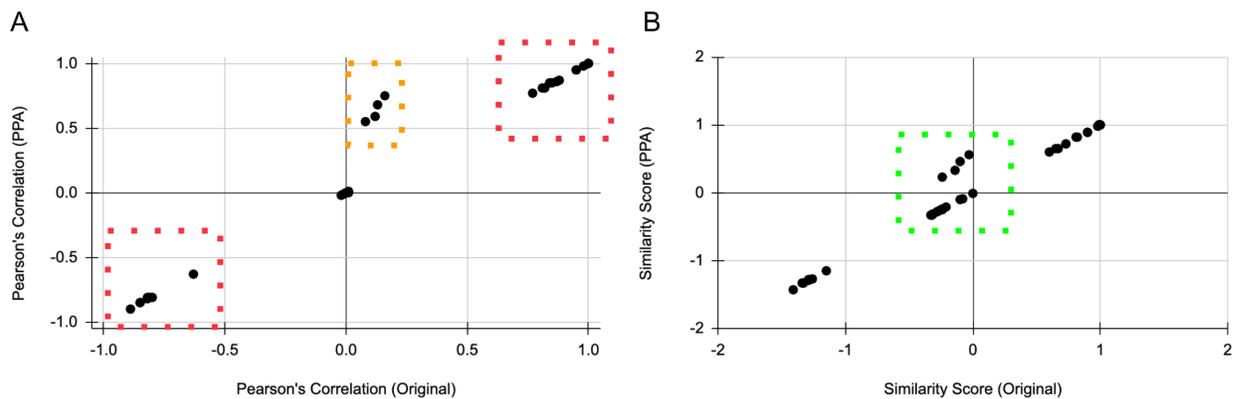

**Fig 8. The accuracy of the piecewise polynomial approximation (PPA) compared to the original data. (A) Comparison of Pearson correlation between raw data and PPA approximation data. (B) Comparison of similarity scores between the original data and the PPA approximation data. PPA compression approximates the original correlation and similarity scores on highly similar and dissimilar tracks.**

## V. Conclusion

Here we built a track comparison tool to reduce costs while maintaining quality of support for users of DSN missions. The track comparison tool provides (1) top 10 similar/dissimilar tracks, (2) anomaly detection within a track, and (3) statistical difference comparison between two given tracks. We also devised a similarity score that combines the three metrics and provides a better contrast when quantifying similarity than Pearson's correlation alone. The preliminary machine learning model showed promising similar/dissimilar track classification performance. We also achieved 1000X time-series data compression to save storage and access data faster for real-time track comparison.

In the next phase, we will increase the number of data sets and perform additional testing to further improve performance prior to planned integration into the track visualizer user interface to support DSN operators and engineers. Multivariate comparisons need to be performed on multiple monitor data items to incorporate dynamic relationships between the various monitor data items. We also need to further filter out tracks of the same type to properly compare multiple tracks (e.g., type of communication, weather). Incorporating expert logic should be considered for more accurate and task-related anomaly detection.

## Acknowledgments

The research was carried out at the Jet Propulsion Laboratory, California Institute of Technology, under a contract with the National Aeronautics and Space Administration (80NM0018D0004). We thank John Mason for his input on the DSN operational use case. We also thank James Montgomery and Lauren Klein for software engineering and data processing/visualization.